\documentclass[runningheads]{llncs}
\usepackage[T1]{fontenc}
\usepackage{tikz}
 \usetikzlibrary{positioning,arrows.meta,shapes}
\usepackage{graphicx}
\usepackage{subcaption}
\begin{document}
\title{From Diagnosis to Redesign: Using Quantitative Ethnography to Improve Multi-Agent LLM Reasoning} %
\titlerunning{Quantitative Ethnography to Improve Multi-Agent LLM Debate}
\author{Vedant Khatri\inst{1}\and
Anthony Cusimano\inst{2} \and
Zachari Swiecki\inst{3} \and
Zhen Xu\inst{4} \and
Xiner Liu\inst{5} \and
Renzhe Yu\inst{4}}
\authorrunning{V. Khatri et al.}
\institute{University of California, Irvine, USA\\
\email{khatriv1@uci.edu}
\and
University of California, Los Angeles, USA\\
\email{anthonycusi@ucla.edu}
\and
Monash University, Australia\\
\email{zach.swiecki@monash.edu}
\and
Columbia University, USA\\
\email{\{zx2393,renzheyu\}@tc.columbia.edu}
\and
University of Pennsylvania, USA\\
\email{xiner@upenn.edu}}
\maketitle

\begin{abstract}
Multi-agent large language model (LLM) systems are designed to improve reasoning by decomposing tasks across multiple agents with specialized functions, but the presence of multiple agents does not inherently guarantee coherent reasoning or outputs that align with task objectives. This paper introduces a quantitative ethnographic (QE) approach for diagnosing and redesigning multi-agent LLM systems based on the discourse produced through agent interactions. We test this approach using automated essay scoring as an example context, applying Epistemic Network Analysis (ENA) to model a five-agent multi-agent debate system and examine differences between debates that produced correct versus incorrect scoring decisions.
Results show that, in the initial system, correct scoring decisions were characterized by rubric-grounded justification, agreement, and elaboration. Incorrect scoring decisions, in contrast,  were characterized by extended proposition-challenge-response exchanges that were less consistently tied to rubric criteria. We then used the findings to revise the agents' prompts. The revised system improved exact scoring accuracy from 27.78\% to 40.28\% and shifted the discourse of incorrect debates toward the rubric-grounded pattern of correct ones, making the two nearly indistinguishable. Based on these results, we argue that QE can support a diagnostic-to-redesign loop for AI reasoning by tracing how patterns of agent interaction relate to system performance, informing prompt redesign, and evaluating whether those redesigns change both outcomes and interaction patterns.

\keywords{Quantitative Ethnography \and Epistemic Network Analysis \and Large Language Models \and Multi-Agent Systems \and Reasoning}
\end{abstract}

\section{Introduction}

Multi-agent large language model (LLM) systems are increasingly used for 
complex reasoning and decision-making tasks, where multiple specialized 
agents collaborate to produce a final output. While these systems can 
outperform single-agent baselines, their reasoning is often 
multi-round and complex, and may involve different types of 
reasoning errors. As a result, it is difficult to understand reasoning 
failure at the task level and to make targeted improvements to the 
agent prompts. Output-level metrics such as accuracy reveal when a system 
fails but not why it fails at the level of the discourse the agents produce \cite{cemri2025,wynn2025}.

To address this challenge, we propose a method that uses Epistemic Network 
Analysis (ENA) \cite{shaffer2017,shaffer2016} to analyze the reasoning 
patterns of multi-agent LLM systems and guide targeted prompt revision. 
ENA represents meaning as patterns of connection among coded discourse 
elements rather than as the frequency of individual elements 
\cite{swiecki2019}, making it well suited to analyze the structured agent 
exchanges produced by multi-agent systems. We use ENA in a 
diagnostic-to-redesign loop: (1) identify the discourse patterns associated 
with successful and unsuccessful system outputs, (2) translate those 
patterns into targeted prompt revisions, and (3) evaluate whether the 
revised system changed both its outputs and the discourse mechanism that 
motivated the redesign.

We test this approach using automated essay scoring as an example context. Essay scoring is a useful test bed because both output accuracy and reasoning quality matter: a holistic essay score is only meaningful if it can be justified in relation to a rubric. Our system is a five-agent Toulmin-structured multi-agent debate (TSMD) in which agents take the roles of Claim, Grounds, Warrant, Rebuttal, and Judge. The five-agent debate produces a complete transcript that can be coded and modeled as discourse.

This work makes three contributions. First, it presents a quantitative ethnographic (QE) methodology for diagnosing the discourse-level failure modes of multi-agent LLM systems rather than evaluating them only by their final 
outputs. Second, it identifies discourse structures associated with accurate 
and inaccurate multi-agent essay scoring in a Toulmin-structured system. 
Third, it demonstrates a diagnostic-to-redesign loop in which ENA findings are 
translated into prompt revisions and evaluated through both scoring 
performance and discourse-structural change. Although the empirical 
demonstration focuses on automated essay scoring, the methodological 
approach is intended to generalize to other multi-agent LLM systems that 
produce deliberative discourse.

\section{Related Work}
\subsection{Multi-Agent LLM Systems and the Problem of Generated Deliberation}

LLM-generated rationales often fail to reflect the actual basis of a model's prediction \cite{pack2024,turpin2023}, motivating designs intended to make reasoning more transparent. Recent work on multi-agent LLM systems has explored whether assigning specialized roles to multiple agents can improve reasoning over single-agent prompting. In this literature, "debate" refers to a structured, sequential exchange in which each agent reads the prior agents' contributions and then adds its own argumentative move before the next agent responds. Du et al. \cite{du2023}  introduced multi-agent debate as a general framework in which two or more LLM instances iteratively critique and revise each other's claims, with the goal of producing more reliable answers than a single agent generating a response in isolation. Liang et al. \cite{liang2024} proposed a multi-agent debate framework in which multiple agents express arguments in a tit-for-tat style and a judge manages the debate process to arrive at a final solution, addressing what they call the "Degeneration-of-Thought" problem: the tendency of a single agent to become unable to generate novel thoughts after establishing confidence in an initial solution.

Beyond the general multi-agent debate framework, recent work has explored role-separated agent designs grounded in normative models of argumentation. Park and Seo \cite{park2025} developed a multi-agent LLM-based debate chatbot for assessing students' critical thinking. Their system organized agent roles around four debate stages derived from Toulmin's argumentation model: establishing evidence, adding a warrant, presenting a rebuttal, and proposing a counter-rebuttal. Their results indicate that role-separated multi-agent designs can substantially improve agreement with human assessment relative to single-agent prompting.

Despite these designs, multi-agent LLM systems often still fail to produce justified reasoning. Cemri et al. \cite{cemri2025} taxonomize their failure modes, finding that role-separated systems frequently produce coordination failures, premature consensus, and unjustified inferences even with explicit role design.  Related work shows that such systems do not reliably outperform simpler prompting strategies \cite{smit2024} and can even shift from correct to incorrect answers across debate rounds \cite{wynn2025}. 

Cemri et al.'s applied a coding-and-counting approach to aggregate failure modes as discrete categorical occurrences in a transcript, which does not attend to how individual agent turns connect to one another to produce the larger argumentative structures. Investigating those interdependencies requires a method that treats the transcript not as a list of discrete utterances but as a network of argumentative relations. This concern matters especially in educational assessment, where a score is only meaningful to the extent that it can be justified. Methods for analyzing the structure of discourse are therefore necessary for distinguishing productive deliberation from plausible-sounding generation.

\subsection{Quantitative Ethnography and Epistemic Network Analysis}

QE is a methodological approach for connecting interpretive coding with quantitative modeling of discourse. In a QE analysis, researchers define a coding scheme grounded in theory, apply that scheme to units of discourse, and then model the patterns of connection among coded elements. ENA \cite{shaffer2017,shaffer2016} is one of the central modeling techniques in QE and represents discourse as networks in which nodes correspond to codes and edges represent co-occurrences of codes within a defined context. The assumption is that meaning is produced through connections among discourse elements, not merely through their frequency.

Swiecki et al. \cite{swiecki2019} argued that collaborative problem solving is an interactive, interdependent, and temporal process that frequency-based methods cannot adequately model because such methods treat individual contributions as isolated and independent events. This finding is directly relevant to multi-agent LLM systems, which produce structured transcripts of role-separated reasoning. If two sets of debates produce similar overall code distributions but different patterns of connection among those codes, ENA can detect the difference where frequency-based methods would not.

Researchers in the QE community have increasingly explored the role of LLMs in qualitative and mixed-methods research workflows. Recent work has examined the use of LLMs for automating qualitative coding \cite{zahid2025}, coding-scheme development \cite{karimov2025}, and multimodal data interpretation \cite{liu2025trajectories}. Because these applications rely heavily on structured prompting to guide LLM responses, attention has turned to how prompt design influences the reasoning processes and results these systems produce \cite{liu2024}. The role of prompting becomes even more consequential in multi-agent LLM systems, where prompts do not simply guide an individual agent's response but shape how agents interact with, challenge, and build upon one another's reasoning as they collectively arrive at decisions.

\subsection{Research Questions}

Despite this growing engagement with LLMs, ENA itself has not been widely applied to AI-generated discourse. At the same time most evaluations of multi-agent LLM systems either rely on output-based performance metrics or on selective qualitative inspection of transcript excerpts. QE and ENA offer a way to address this gap by representing the entire generated transcript as network structures that allow systematic comparisons across agent configurations and prompting conditions.
We treat the generated transcript of a multi-agent LLM system as analyzable discourse, use ENA to identify discourse structures associated with successful and unsuccessful outputs, and then use that diagnosis to revise the system. This is a diagnostic-to-redesign loop, in which the analytic method is not only describing the system but also providing targets for redesign. 

Using ENA in this way changes the kind of claims that can be made about system improvement. A redesign motivated by a discourse-level diagnosis is testable in two ways: it should change output performance, and it should change the discourse mechanism that was identified as the problem. The second test is essential because output improvement alone does not establish that the intervention worked at the intended level.

In sum, we test the diagnostic-to-redesign loop via the following research questions:
\begin{description}
    \item[RQ1.] What differences in discourse network structure distinguish debates that produce accurate scores from those that produce inaccurate scores in a multi-agent essay-scoring system?
    \item[RQ2.] Can discourse patterns identified through ENA inform revisions that improve scoring accuracy?
    \item[RQ3.] How does the discourse network structure of the redesigned system compare with the network structure associated with accurate scoring in the initial system?
\end{description}

\section{Methods}

\subsection{Data and Sample}

We used essays from the Automated Student Assessment Prize 2.0 (ASAP 2.0) corpus \cite{crossley2025}, a benchmark dataset of argumentative essays written by U.S. secondary students. Each essay was scored on a six-point holistic scale ranging from 1 (very little or no mastery of writing) to 6 (clear and consistent mastery), with intermediate bands marking progressively stronger development of point of view, evidence use, organization, and language control. Gold-standard scores were assigned by trained expert raters following the corpus's published scoring procedure \cite{crossley2025}. From this corpus, we drew a stratified sample of 72 essays, with twelve essays from each score level, ensuring the evaluation represented the full scoring range rather than over-representing middle-score essays.

\subsection{Multi-Agent System Architecture}

We adopt Toulmin's \cite{toulmin2003} argumentation model---which describes argument as a structured relationship among a claim, the grounds supporting it, the warrant licensing the inferential move from grounds to claim, and a rebuttal identifying conditions under which the claim would not hold---as the theoretical frame for both our system architecture and our discourse analysis. %
Park and Seo \cite{park2025} demonstrated that mapping Toulmin's components onto separate LLM agents substantially improves agreement with human evaluators on assessment tasks, while Liang et al.\cite{liang2024} showed that adding a dedicated Judge Agent to synthesize a multi-agent debate addresses the Degeneration-of-Thought problem that limits single-agent revision. Our Toulmin-structured multi-agent debate (TSMD) system combines these contributions: four agents take the Claim, Grounds, Warrant, and Rebuttal roles, and a fifth Judge Agent produces the final score.

\begin{figure}[t]
\centering
\begin{tikzpicture}[
  node distance=0.25cm,
  agent/.style={rectangle, draw=black!70, rounded corners=3pt, 
                minimum width=7cm, minimum height=0.6cm, 
                align=center, font=\footnotesize\sffamily, fill=blue!5},
  io/.style={rectangle, draw=black!50, dashed, 
             minimum width=4.5cm, minimum height=0.5cm, 
             align=center, font=\footnotesize\sffamily\itshape, fill=gray!10},
  arrow/.style={->, >=stealth, thick, black!70}
]
\node[io] (input) {Essay text + ASAP 2.0 rubric};
\node[agent, below=of input] (claim) {\textbf{Claim Agent}: proposes initial score (1--6)};
\node[agent, below=of claim] (grounds) {\textbf{Grounds Agent}: identifies textual evidence};
\node[agent, below=of grounds] (warrant) {\textbf{Warrant Agent}: maps evidence to rubric criteria};
\node[agent, below=of warrant] (rebuttal) {\textbf{Rebuttal Agent}: challenges argument, considers revision};
\node[agent, below=of rebuttal] (judge) {\textbf{Judge Agent}: synthesizes debate, commits to final score};
\node[io, below=of judge] (output) {Final holistic score};
\draw[arrow] (input) -- (claim);
\draw[arrow] (claim) -- (grounds);
\draw[arrow] (grounds) -- (warrant);
\draw[arrow] (warrant) -- (rebuttal);
\draw[arrow] (rebuttal) -- (judge);
\draw[arrow] (judge) -- (output);
\end{tikzpicture}
\caption{TSMD architecture. Each essay is processed sequentially through five agents corresponding to Toulmin's argumentative components and a final Judge Agent. Every agent receives the essay text, the ASAP 2.0 rubric, and the full transcript of prior agent turns before producing its contribution.}
\label{fig:tsmd_architecture}
\end{figure}
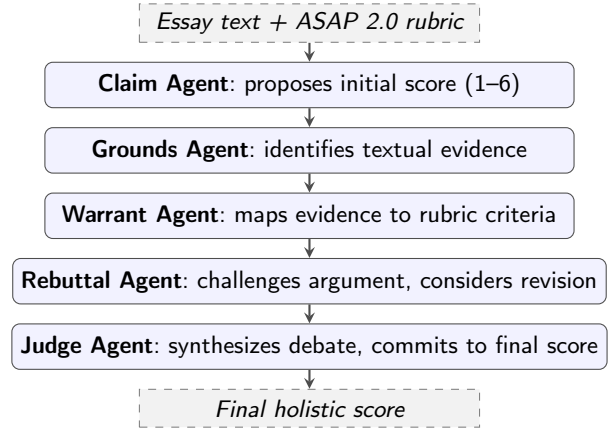

\paragraph{Agent Role Assignment.}
Each essay was scored through a sequence of five agent turns; Figure~\ref{fig:tsmd_architecture} shows the architecture and the role of each agent.

\paragraph{Debate Process.}

At each turn, the active agent received a prompt structured into two parts. Following prior work \cite{liang2024,park2025}, we used a two-part prompt structure: the system prompt contained role-specific instructions defining the agent's argumentative function (Claim, Grounds, Warrant, Rebuttal, or Judge) and the ASAP 2.0 scoring rubric, while the user prompt contained the essay text and the complete transcript of prior agent turns. In both experimental rounds, the rubric was provided as part of every agent's system prompt; the intervention in the revised round (see Section~4.6) did not add the rubric but rather changed the system-prompt instructions to require explicit evidence-to-rubric mapping at each agent's turn. Full prompt templates for the agents in both rounds are provided in Appendix~\ref{app:prompts}.

\paragraph{Model and Sampling.}
All five agents were instantiated with Qwen 2.5 72B Instruct \cite{yang2024}, an open-weight large language model selected for its strong instruction-following performance and for supporting reproducibility without requiring proprietary API access. Each agent turn was sampled with a temperature of 0.3 and a maximum of 400 new tokens per turn. The low temperature was chosen to produce stable outputs appropriate for a scoring task with enough variation for agents to produce distinct argumentative moves. Identical sampling parameters were used across all five agents and across both experimental rounds.

\paragraph{Design Contributions.}
Two aspects of this architecture differ from prior multi-agent debate systems. First, the Rebuttal Agent uses a bidirectional design that can argue for either an upward or downward score revision, or agree if no revision is warranted, whereas prior rebuttal designs typically instruct an agent to disagree with the proposing agent in a single direction. Second, the Claim Agent prompt includes few-shot examples for each score level on the rubric, rather than a single generic prompt.

\subsection{Epistemic Network Analysis}

We coded each agent turn using the Conversational Argument Coding Scheme (CACS) \cite{canary2010}, which characterizes the communicative function of utterances in argumentative discourse. We selected CACS because it was designed to capture the argumentative moves of multi-party conversational argument: assertions, supports, objections, and responses that arise as participants build and challenge claims through interaction. Our system produces analogous structured exchanges among role-separated agents, making CACS a natural fit for characterizing the argumentative function of each agent's turn.

CACS organizes argumentative moves into six broader categories. Four of these categories were used in our coding: Starting Points (Assertion, Proposition); Developing Points (Elaboration, Amplification, Justification); Convergence Markers (Agreement, Acknowledgement); and Prompters (Objection, Challenge, Response). The remaining two categories (Delimiters and Non-Arguments) include four codes (Frame, Forestall/Secure, Forestall/Refute, Non-Argument) that did not occur in our data and are not reported here.

We applied CACS deductively, using the code definitions and category structure as published in Canary and Seibold's revised scheme without modification. The coders did not introduce new codes or merge existing ones during the coding process; rather, the analysis examined which of the established CACS categories were used in the multi-agent debate corpus and which were not. Table~\ref{tab:cacs} lists the ten codes, their definitions from Canary and Seibold \cite{canary2010}, and example utterances drawn from the agent transcripts.

\begin{table}[!htb]
\centering
\caption{CACS codes used in this study, with definitions and example agent utterances. Code definitions are taken from Canary and Seibold \cite{canary2010}, Table 2 (revised version).}
\label{tab:cacs}
\scriptsize
\setlength{\tabcolsep}{4pt}
\renewcommand{\arraystretch}{1.15}
\begin{tabular}{p{2.4cm}p{4.2cm}p{5.0cm}}
\hline
\textbf{Code} & \textbf{Definition} & \textbf{Example utterance} \\
\hline
Assertion & Statements of belief or opinion. & ``This essay seems to deserve a score of 2.'' [Claim Agent] \\
Proposition & Statements that call for discussion or action. & ``The current score of 4 is reasonable, but I would lean towards a 3.'' [Rebuttal Agent] \\
Elaboration & Statements that support other statements by providing evidence or clarification. & ``The essay presents a vague and underdeveloped argument against the idea that aliens created the Face on Mars.'' [Grounds Agent] \\
Amplification & Explicit inferential statements. & ``These factors suggest the essay has a point of view but is weakly supported, aligning with a score of 3.'' [Grounds Agent] \\
Justification & Statements that offer norms, values, or rules of logic to support the validity of other statements. & ``The essay fits the rubric description of a 2 because it has a vague, very limited, or poorly supported point of view.'' [Rebuttal Agent] \\
Agreement & Statements that show agreement. & ``The proposed score of 2 is justified.'' [Warrant Agent] \\
Acknowledgement & Messages indicating recognition and/or understanding, but not agreement to another's point. & ``While the essay includes relevant evidence, it falls short of the consistent mastery required for a 5.'' [Rebuttal Agent] \\
Objection & Statements that deny the truth or accuracy of another statement. & ``Counter-proposal: I argue for a score of 3.'' [Rebuttal Agent] \\
Challenge & Messages that present problems, questions, or reservations that must be addressed to reach agreement. & ``However, the development of ideas is somewhat uneven, and the organization could be tighter.'' [Claim Agent] \\
Response & Statements that support other statements that have been explicitly refuted. & ``Despite these issues, the essay demonstrates adequate mastery but has room for improvement in coherence and language control.'' [Grounds Agent] \\
\hline
\end{tabular}
\end{table}

Each agent turn was coded for the presence or absence of every code in the scheme: a code was marked 1 if it appeared in the turn and 0 otherwise, allowing multiple codes per turn. This turn-level unit served as the line for subsequent ENA modeling. The two coders---a graduate and an undergraduate researcher in computer science, both actively involved in learning analytics and education research---jointly coded a calibration set of pilot agent turns to establish shared interpretations of each code. They then coded every agent turn independently following the negotiated coding methodology of Garrison et al. \cite{garrison2006}, with disagreements resolved through social moderation to produce a final consensus code.

We used ENA \cite{shaffer2017} to model and compare the relationship and co-occurrence of the generated argumentation among agents for each essay. Each essay-level debate was also treated as a separate conversation/horizon, such that co-occurrences among codes were calculated only within the discussion associated with a single essay and not across debates for different ones. We selected an infinite moving window size because the argumentative structure of the multi-agent system was distributed across the full debate rather than localized to adjacent turns. As agents frequently referenced, challenged, and elaborated on claims introduced earlier in the discussion, meaningful relationships among coded moves could occur anywhere within the entire essay-level conversation.

\subsection{Prompt Revision}

We used the epistemic networks to identify discourse patterns that distinguished accurate from inaccurate debates, and revised the agent prompts based on the results of RQ1 (see Section~4.1). The analysis indicated that the system's primary failure mode was a lack of rubric-grounded warranting: agents could discuss the essay and propose scores, but they did not consistently anchor those moves in specific rubric criteria.

We revised the agent prompts to address this deficit at the discourse level, while keeping the rubric, sampling parameters, and architecture identical across rounds. The Warrant Agent received the most substantial revision, now required to name the relevant rubric criterion before evaluating whether the essay meets that criterion---directly targeting the Justification deficit identified in RQ1. The Grounds and Warrant Agents were also prohibited from proposing new scores, restricting score-proposal moves to the Claim and Rebuttal Agents in order to reduce the unanchored Proposition behavior observed in inaccurate debates. Appendix~\ref{app:prompts} summarizes the prompt changes by agent role.

All other system parameters remained constant. The same 72 essays were then re-scored using the revised prompts, producing a second set of 72 debate transcripts. These transcripts were coded using the same CACS and social moderation procedure.

\subsection{Analytical Pipeline}

For RQ1, we constructed epistemic networks from the debate transcripts generated before the prompt revision and used means rotation to align the first dimension with the greatest separation between the accurate and inaccurate scoring groups. The grouping variable was prediction accuracy: debates were classified as accurate if the Judge Agent's final score exactly matched the gold-standard human score and inaccurate otherwise. Group differences were tested using a two-tailed Welch's $t$-test, because it accommodates the unequal group sizes and variances present in this comparison (20 accurate vs.\ 52 inaccurate debates). The two-tailed specification was chosen because we did not hold a directional hypothesis about the network-structure differences. Cohen's $d$ is reported as the matched parametric effect size.

For RQ2, we compared exact-match scoring accuracy before and after the prompt revision. Exact accuracy was defined as the proportion of essays for which the Judge Agent's final score exactly matched the gold-standard human score. Because the same 72 essays were scored in both rounds, we used a one-tailed McNemar's chi-square test to evaluate whether more essays changed from inaccurate to accurate than from accurate to inaccurate. The one-tailed specification reflected our directional hypothesis that the prompt revision would improve accuracy, and the paired-comparison structure---holding essay sample, agent architecture, and model settings constant across rounds---motivated McNemar's test rather than an independent-samples test.

For RQ3, we modeled the transcripts generated after the prompt revision using the same ENA parameters as in RQ1 and projected the resulting essay-level networks into the ENA space constructed in RQ1. This allowed direct comparison of accurate and inaccurate revised-round debates along the same MR1 dimension that distinguished accurate from inaccurate debates in the initial round. As with RQ1, group differences on MR1 were tested using a two-tailed Welch's $t$-test (29 accurate vs.\ 43 inaccurate debates), with Cohen's $d$ reported as the effect size. The key indicator of redesign success was the reduction in centroid separation between accurate and inaccurate groups, evaluated against the initial-round model.

\section{Results}

\subsection{Accurate Debates Were More Rubric-Grounded (RQ1)}

RQ1 asked whether accurate and inaccurate debates differed in their argumentative discourse network structure. In the initial round, the system produced a scoring accuracy of 27.78\% with exact-match scores for 20 of the 72 essays and non-exact scores for 52 essays. We compared the ENA networks of these accurate and inaccurate debates.

The means-rotated ENA model projected the 72 essay-level networks into a two-dimensional space in which MR1 captured the discourse dimension that best separated accurate from inaccurate debates. A two-tailed Welch's $t$-test on MR1 showed a statistically significant difference between the two groups, $t(37.48) = -4.69$, $p < .001$, $d = 1.19$. The accurate group centroid was located at MR1 = -0.21 ($SD = 0.23$), and the inaccurate group centroid was located at MR1 = 0.08 ($SD = 0.25$).

\begin{figure}[t]
\centering
\includegraphics[width=0.9\textwidth]{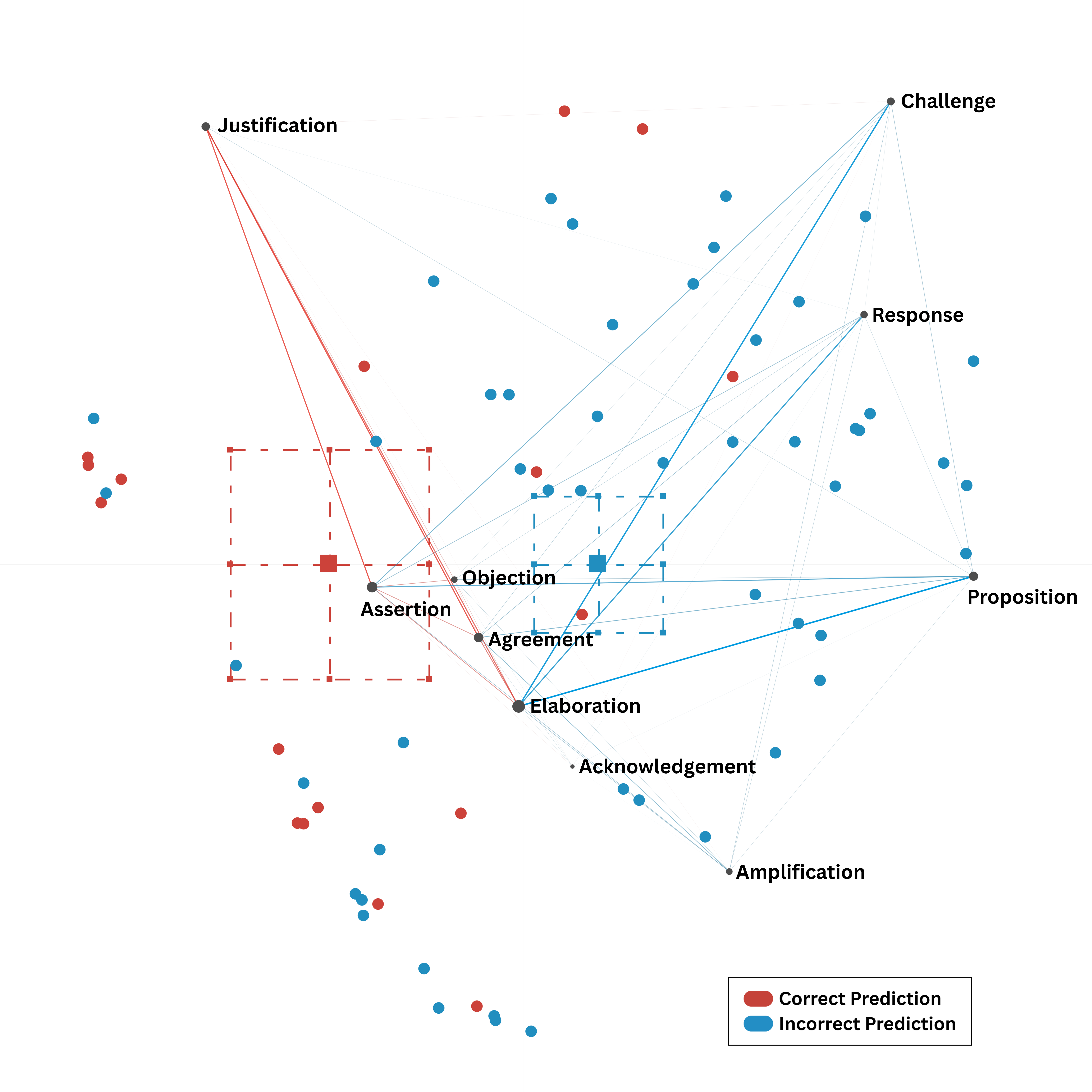}
\caption{ENA mean networks for accurate and inaccurate debates in the initial round.}
\label{fig:rq1_networks}
\end{figure}

As shown in Figure~\ref{fig:rq1_networks}, the mean network for accurate debates showed strong connections among Justification, Agreement, Elaboration, and Assertion. We interpret this pattern as rubric-grounded convergence, as agents elaborated essay features, justified their relevance using rubric criteria, and aligned around a score interpretation. The mean network for inaccurate debates showed stronger connections among Proposition, Challenge, and Response. We interpret this pattern as unanchored contestation, as agents proposed, challenged, and responded to scores without consistently connecting those moves to the rubric.

\subsection{ENA-Informed Prompt Revision Improved Exact Accuracy (RQ2)}

RQ2 asked whether  the findings from RQ1 could inform a prompt revision that improved scoring accuracy. Based on prior findings, we revised the agents' prompts to require explicit rubric citation and evidence-to-criterion mapping. The same 72 essays were then re-scored under the revised prompts.

Exact accuracy increased from 27.78\% (20/72) in the initial round to 
40.28\% (29/72) in the revised round, a relative improvement of 
approximately 45\%. A McNemar chi-square test indicated that this paired 
improvement was statistically significant, $\chi^2(1) = 3.24$, one-tailed 
$p = .0359$.

The paired structure of the data also allows a more detailed breakdown of 
the accuracy change. Of the 20 essays scored correctly in the initial round, 
12 remained correct after revision and 8 became incorrect. Of the 52 essays 
scored incorrectly in the initial round, 17 became correct after revision 
and 35 remained incorrect. The net change of plus 9 essays moving from 
incorrect to correct corresponds to the 12.5 percentage point improvement 
reported above.

The discourse distribution also changed substantially across rounds. 
Justification increased from 52 turns in the initial round to 225 turns in 
the revised round, a more than four-fold increase. Agreement also increased 
(from 139 to 159 turns), while Objection and Response decreased (Objection 
from 57 to 12 turns; Response from 43 to 5 turns). These shifts are 
consistent with the design intent of the prompt revision: the revised 
agents produced substantially more rubric-grounded justification and 
proportionally less unanchored proposal-and-objection exchange.

\subsection{Revised Prompts Reduced the Discourse-Level Separation Between Groups (RQ3)}

RQ3 asked whether the prompt revision produced the discourse network structure that more closely resembled that associated with accurate scoring in RQ1. This question is important because improved accuracy alone does not establish that the intervention worked at the discourse level. The system could have become more accurate for reasons unrelated to the discourse pattern identified in RQ1. 
To examine this, we accumulated co-occurrences in the revised-round corpus using the same CACS codes, turn-level lines, infinite window, and normalization parameters as the initial-round model, and projected the resulting essay-level networks into the ENA space created in RQ1.

\begin{figure}[t]
\centering
\includegraphics[width=0.95\textwidth]{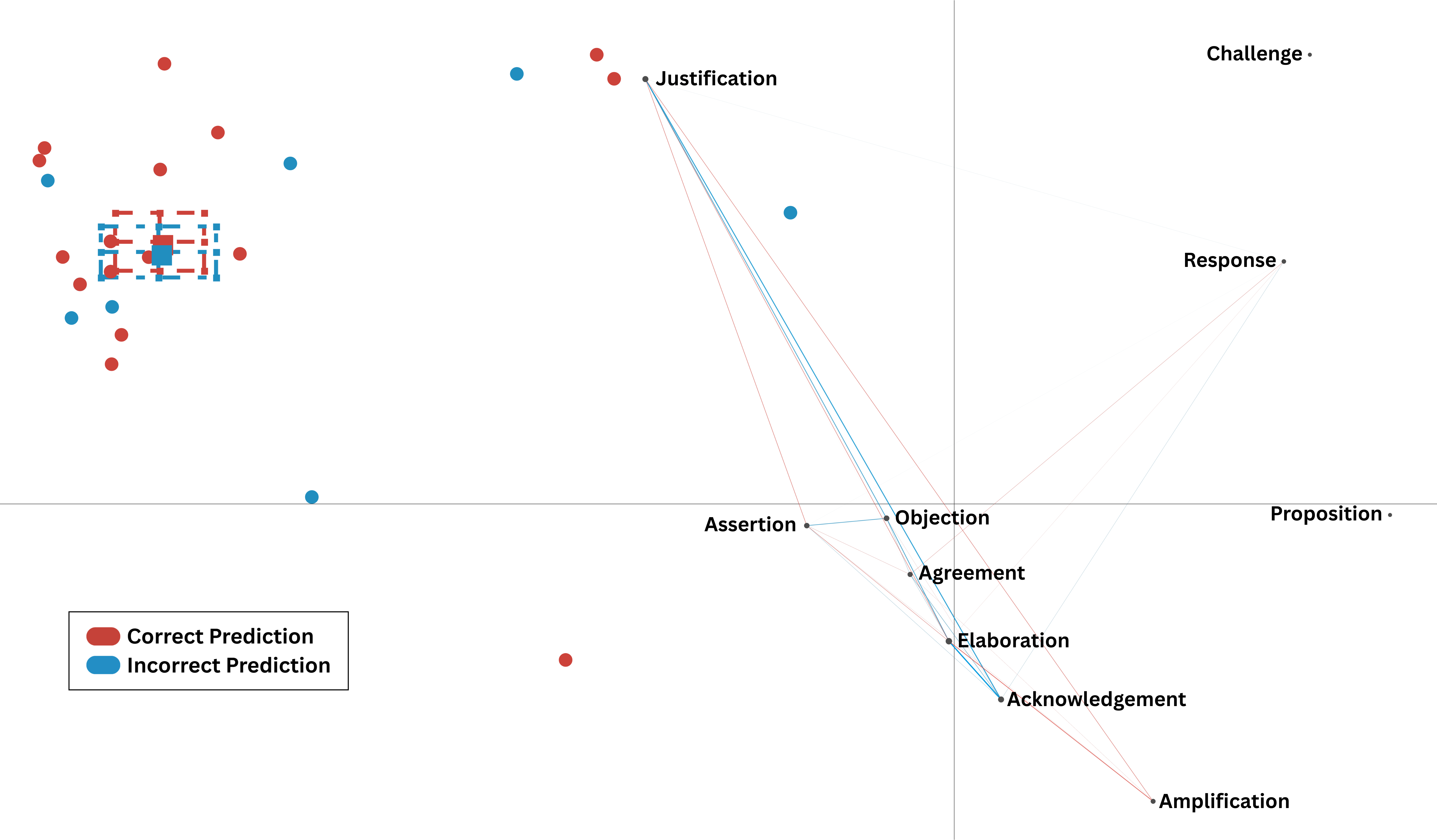}
\caption{Revised-round ENA model comparing accurate and inaccurate debates.}
\label{fig:rq3_networks}
\end{figure}

From this perspective, the key result is not simply whether the revised-round accurate and inaccurate groups differed significantly. Rather, the key question is whether the revised prompts reduced the discourse-level separation that characterized the initial round. Unlike the initial-round model, the revised-round ENA model did not show a statistically significant difference between accurate and inaccurate debates (Figure~\ref{fig:rq3_networks}). On the first projected dimension, a two-tailed Welch's $t$-test returned $t(58.30) = -0.03$, $p = .97$, $d = 0.01$. The accurate group centroid was located at V1 = -0.42 ($SD = 0.08$, $N = 29$), and the inaccurate group centroid was located at V1 = -0.42 ($SD = 0.08$, $N = 43$). This corresponds to a centroid separation of approximately 0 in the revised-round ENA space. By contrast, the initial-round model showed a much larger separation of 0.29. This convergence could also reflect prompts making all debates sound uniformly rubric-grounded regardless of score correctness, rather than discourse explaining accuracy.

\section{Discussion and Conclusion}

This study tested whether ENA could identify discourse structures associated with accurate and inaccurate scoring in multi-agent LLM debates and inform a targeted redesign of the system. RQ1 showed a substantial difference in discourse network structure between debates that produced accurate scores and those that produced inaccurate scores: accurate debates were organized around rubric-grounded justification, while inaccurate debates unfolded as cycles of proposition, challenge, and response without consistent rubric grounding. RQ2 showed that an ENA-informed prompt revision targeted at this discourse-level deficit produced a statistically significant improvement in exact scoring accuracy. RQ3 showed that, after the revision, accurate and inaccurate debates contracted into highly similar regions of the projected discourse space, providing converging evidence that the redesign worked at the level it was intended to address.

These findings contribute to two literatures. For multi-agent LLM systems, our RQ1 result complicates the claim that role-separated multi-agent designs improve assessment accuracy. Prior work has evaluated such systems primarily by output metrics such as scoring accuracy or agreement with human raters \cite{liang2024,park2025}. We show that role separation does not, on its own, produce the evidence-to-rubric reasoning that the design is meant to encourage: in the initial round, inaccurate debates still consisted of proposal-and-challenge cycles without rubric grounding. The accuracy benefits of role-separated multi-agent designs depend on whether agents actually produce the kind of reasoning the design is supposed to elicit. The RQ2 result then offers a way to address this gap by translating output-level failure into a discourse-level target that can be acted on. The revised prompts in our system were not chosen by intuition or generic prompt-engineering practice but derived directly from the discourse pattern that distinguished accurate from inaccurate debates. Park and Seo \cite{park2025} showed that Toulmin-structured agent roles can improve agreement with human raters; our results extend this by showing that explicitly requiring those roles to cite rubric criteria in the prompt can produce further, discourse-grounded gains while preserving a deliberative structure that researchers can inspect.

For QE, our study examines a relatively new use of ENA, not only as an analytic tool but also as a source of evidence for system redesign.
 Most ENA studies have analyzed human discourse, including collaborative problem solving \cite{swiecki2019}, online learning communities, and classroom interaction. We extend ENA to LLM-generated discourse, treating the transcript of a multi-agent debate as analyzable structure rather than as informal explanation, and we use it not only to describe discourse but to actively guide system redesign. Our findings are also consistent with the argument of Swiecki et al. \cite{swiecki2019} that frequency-based methods miss the interactive, interdependent structure ENA can capture: in the initial round, the two accuracy groups overlapped substantially in which CACS codes appeared but differed sharply in how those codes connected. This is precisely the kind of finding that frequency-based discourse analysis would miss, and it underscores the methodological value of treating AI-generated discourse as networked rather than as a list of moves.

Our study includes several limitations that open promising directions for future work. While our diagnostic-to-redesign loop produced a statistically significant accuracy improvement (27.78\% to 40.28\%) and meaningful discourse-level convergence, the absolute gain was modest and the revised system remained inaccurate on a majority of essays, suggesting that the discourse-level deficit identified in RQ1 is not the only mechanism driving inaccuracy. Other factors not captured by CACS coding---such as rubric interpretation quality, evidence selection, and inter-agent deliberation depth---likely contribute to remaining errors, and the present study was scoped to a single domain (essay scoring) using one open-weight model (Qwen 2.5 72B Instruct \cite{yang2024}). We also did not include a single-agent baseline or an ablation isolating which prompt changes mattered, which future work should address. The most immediate extension is to apply the diagnostic-to-redesign loop iteratively, identifying the next discourse-level mechanism with ENA and targeting it with a second round of revision; additional directions include extending the architecture to allow multiple debate rounds, requiring the Judge Agent to explicitly compare competing rubric mappings, providing rubric-anchored exemplars at each score band, and replicating the approach across other reasoning contexts and other LLM families to clarify how stable the diagnosed discourse structures are across contexts and models.

In our study, the diagnose-design-re-evaluate loop transformed an output-level problem---inaccurate essay scores---into a discourse-level target (insufficient rubric-grounded warranting), operationalized that target through prompt revision, and evaluated whether the discourse changed in the predicted direction. This loop is not specific to essay scoring. Any multi-agent LLM system that produces deliberative discourse could in principle be analyzed this way, including clinical decision-support agents, legal-reasoning systems, and collaborative planning tools. The codes and success criteria would differ, but the methodological logic remains the same: identify discourse structures associated with successful outcomes, translate them into design targets, and evaluate whether the intervention changes both performance and discourse. As multi-agent LLM systems are deployed in education and other settings where reasoning quality matters, the contribution of work like this is not a single accuracy gain but a methodology for ensuring that what these systems produce can be inspected, interpreted, and improved at the level of the discourse they generate.
\begin{credits}
\subsubsection{\ackname}
We thank David Williamson Shaffer for his guidance on this work, and we are grateful to Sandy Irani and Shannon Alfaro for their support. This work was supported in part by the Summer Undergraduate Research Program (SURP) through the Undergraduate Research Opportunities Program (UROP) at the University of California, Irvine.
\end{credits}

\appendix
\section{Appendix: Agent Prompts}
\label{app:prompts}
The full original and revised prompts are provided in a supplementary repository:
\newline
https://github.com/AnthonyCusi/ICQE-appendix


\begin{thebibliography}{20}

\bibitem{liang2024}
Liang, T., He, Z., Jiao, W., Wang, X., Wang, Y., Wang, R., Yang, Y., Shi, S., Tu, Z.: Encouraging divergent thinking in large language models through multi-agent debate. In: Proceedings of the 2024 Conference on Empirical Methods in Natural Language Processing (EMNLP), pp. 17889--17904. Association for Computational Linguistics, Miami, Florida, USA (2024). \doi{10.18653/v1/2024.emnlp-main.992}

\bibitem{smit2024}
Smit, A.P., Grinsztajn, N., Duckworth, P., Barrett, T.D., Pretorius, A.: Should we be going MAD? A look at multi-agent debate strategies for LLMs. In: Proceedings of the 41st International Conference on Machine Learning (ICML). PMLR (2024). arXiv:2311.17371

\bibitem{park2025}
Park, B., Seo, K.: Assessing critical thinking through a multi-agent LLM-based debate chatbot. In: Extended Abstracts of the CHI Conference on Human Factors in Computing Systems (CHI EA '25), Yokohama, Japan. ACM, New York (2025). \doi{10.1145/3706599.3721207}

\bibitem{toulmin2003}
Toulmin, S.E.: The Uses of Argument, updated edn. Cambridge University Press, Cambridge (2003) [Original work published 1958]

\bibitem{shaffer2017}
Shaffer, D.W.: Quantitative Ethnography. Cathcart Press, Madison (2017)

\bibitem{shaffer2016}
Shaffer, D.W., Collier, W., Ruis, A.R.: A tutorial on epistemic network analysis: analyzing the structure of connections in cognitive, social, and interaction data. J. Learn. Anal. 3(3), 9--45 (2016). \doi{10.18608/jla.2016.33.3}

\bibitem{swiecki2019}
Swiecki, Z., Ruis, A.R., Farrell, C., Shaffer, D.W.: Assessing individual contributions to collaborative problem solving: a network analysis approach. Comput. Hum. Behav. 104, 105876 (2019). \doi{10.1016/j.chb.2019.01.009}

\bibitem{yang2024}
Yang, A., Yang, B., Hui, B., Zheng, B., Yu, B., Li, C., Liu, D., Huang, F., Wei, H., Lin, H., et al.: Qwen2.5 Technical Report. arXiv preprint arXiv:2412.15115 (2024)

\bibitem{canary2010}
Canary, D.J., Seibold, D.R.: Origins and development of the conversational argument coding scheme. Commun. Methods Meas. 4(1--2), 7--26 (2010). \doi{10.1080/19312451003680459}

\bibitem{garrison2006}
Garrison, D.R., Cleveland-Innes, M., Koole, M., Kappelman, J.: Revisiting methodological issues in transcript analysis: negotiated coding and reliability. Internet High. Educ. 9(1), 1--8 (2006)





\bibitem{pack2024}
Pack, A., Barrett, A., Escalante, J.: Large language models and automated essay scoring of English language learner writing: insights into validity and reliability. Comput. Educ. Artif. Intell. 6, 100234 (2024). \doi{10.1016/j.caeai.2024.100234}

\bibitem{turpin2023}
Turpin, M., Michael, J., Perez, E., Bowman, S.R.: Language models don't always say what they think: unfaithful explanations in chain-of-thought prompting. In: Advances in Neural Information Processing Systems 36, pp. 74952--74965 (2023)

\bibitem{cemri2025}
Cemri, M., Pan, M.Z., Yang, S., Agrawal, L.A., Chopra, B., Tiwari, R., Keutzer, K., Parameswaran, A., Klein, D., Ramchandran, K., Zaharia, M., Gonzalez, J.E., Stoica, I.: Why do multi-agent LLM systems fail? In: Advances in Neural Information Processing Systems 38, Datasets and Benchmarks Track (2025)

\bibitem{wynn2025}
Wynn, A., Satija, H., Hadfield, G.: Talk isn't always cheap: understanding failure modes in multi-agent debate. In: ICML 2025 Workshop on Multi-Agent Systems (MAS). arXiv:2509.05396 (2025). \doi{10.48550/arXiv.2509.05396}


\bibitem{crossley2025}
Crossley, S.A., Baffour, P., Burleigh, L., King, J.: A large-scale corpus for assessing source-based writing quality: ASAP 2.0. Assess. Writ. 65, 100954 (2025). \doi{10.1016/j.asw.2025.100954}

\bibitem{du2023}
Du, Y., Li, S., Torralba, A., Tenenbaum, J.B., Mordatch, I.: Improving factuality and reasoning in language models through multiagent debate. arXiv preprint arXiv:2305.14325 (2023)

\bibitem{zahid2025}
Zahid, O., Hsiang-Pan, J., Arastoopour Irgens, G., Behboudi, A., Lane, A.C.: Of humans and machines: evaluating the efficacy of GPT-4 in coding discourse data. In: International Conference on Quantitative Ethnography, pp. 117--132 (2025)

\bibitem{karimov2025}
Karimov, A., Saarela, M., Liu, X., Wei, Z., Zambrano, A.F., Barany, A., K\"arkk\"ainen, T.: ChatGPT-assisted codebook design for learning analytics datasets in multiple languages: a case study. In: International Conference on Quantitative Ethnography, pp. 177--192 (2025)

\bibitem{liu2025trajectories}
Liu, X., Zhou, Y., Ocumpaugh, J., Barany, A., Zambrano, A.F., Wei, Z., Giordano, C.: Not all who wander are lost: trailblazing trajectories in a Minecraft-based learning environment. In: International Conference on Quantitative Ethnography, pp. 336--352 (2025)

\bibitem{liu2024}
Liu, X., Zhang, J., Barany, A., Pankiewicz, M., Baker, R.S.: Assessing the potential and limits of large language models in qualitative coding. In: International Conference on Quantitative Ethnography, pp. 89--103. Springer Nature Switzerland, Cham (2024)

\end{thebibliography}
\end{document}